\documentclass{article}

\usepackage{PRIMEarxiv}

\usepackage[utf8]{inputenc} 
\usepackage[T1]{fontenc}    
\usepackage{hyperref}       
\usepackage{url}            
\usepackage{booktabs}       
\usepackage{amsfonts}       
\usepackage{nicefrac}       
\usepackage{microtype}      
\usepackage{lipsum}
\usepackage{fancyhdr}       
\usepackage{graphicx}       
\graphicspath{{media/}}     
\usepackage{multirow}

\title{Improving Zero-Shot Action Recognition using Human Instruction with Text Description
}

\author{
  Nan Wu \\
  Graduate School of Science and Engineering \\
  Chiba University \\
  Chiba, Japan\\
  \texttt{gonan@chiba-u.jp} \\
   \And
  Hiroshi Kera, Kazuhiko Kawamoto \\
  Graduate School of Engineering \\
  Chiba University \\
  Chiba, Japan\\
  \texttt{kera@chiba-u.jp, kawa@faculty.chiba-u.jp} \\
}

\begin{document}
\maketitle

\begin{abstract}
Zero-shot action recognition, which recognizes actions in videos without having received any training examples, is gaining wide attention considering it can save labor costs and training time.
Nevertheless, the performance of zero-shot learning is still 
unsatisfactory, which limits its practical application. 
To solve this problem, this study proposes a framework to improve zero-shot action recognition using human instructions with text descriptions.
The proposed framework manually describes video contents, which incurs some labor costs; in many situations, the labor costs are worth it.
We manually annotate text features for each action, which can be a word, phrase, or sentence.
Then by computing the matching degrees between the video and all text features, we can predict the class of the video.
Furthermore, the proposed model can also be combined with other models to improve its accuracy. 
In addition, our model can be continuously optimized to improve the accuracy by repeating human instructions.
The results with UCF101 and HMDB51 showed that our model achieved the best accuracy and improved the accuracies of other models.
\end{abstract}


\section{Introduction}\label{sec1}

Machine learning (ML) has advanced significantly in recent years, resulting in improved accuracy in ML models. However, the increasing size of training datasets has created new challenges. While larger datasets can improve model accuracy, they require increased manual labor for labeling, making it difficult for ordinary people to create their own datasets. Furthermore, large datasets lead to longer training times, especially with video datasets that contain vast amounts of data. In addition, retraining the model is also time-consuming. These challenges make model training a time-intensive process, and the inability to find suitable datasets can be discouraging.

To solve this problem, researchers have studied zero-shot learning \cite{awa,lampert2013attribute}, which allows models to classify objects even if they have never encountered a certain class of objects. This method does not require large amounts of data for training the model. Therefore, it can save labor costs and training time. New classes can be added to the model without retraining it. 
Nevertheless,
the performance of zero-shot learning is still unsatisfactory, which limits its practical application. 
For example, 
the accuracies of the zero-shot action recognition
methods \cite{qin2017zero,zhu2018towards,tsgcn}
are 15.1\%, 17.5\%, and 33.4\%, respectively
for the UCF101 \cite{ucf101} dataset with 50 unseen classes.

This study proposes a framework for improving zero-shot action recognition using human instructions with text descriptions.
In the human instructions, users describe texts, which can be a word, phrase, or sentence, by illustrating
the content of a given video and feed them into the model.

Although the human instructions incurs some labor costs, 
in many situations, the costs are worth it. 
We developed a video-text matching model (VTMM) to fuse the video and text features after human
instructions.
The VTMM provides the matching degree between these two features. 
For example, 
if we describe ``in the water'' for a video demonstrating swimming,
the matching degree takes a high value. In contrast, 
if we describe ``a hamburger in hand'' for the same video,
it takes a low value.
Furthermore, we annotate some common features.
For example, we annotate several features with text for the action “Billiards,” such as (i) ``A man holds a stick in his hand,'' (ii) ``A man is standing at a table,'' and (iii) ``There are many balls on the table.''
We also annotate some common short descriptions, such as \emph{indoor}, \emph{action is slow}, and \emph{standing}.
The matching degree helps the recognition model to determine the video class.

VTMM can be used alone without relying on any other model. Each action is annotated with one or more features. After annotating the features, we put the video and all text features into the model, knowing which features the video can match with. Based on the action classes to which the features belong, we can calculate the probability that the video belongs to each action class.
In addition, our model can be appended to any other model to improve accuracy without retraining the model. 
All we have to do is add features to the classes that need to be optimized without modifying other classes.
For example, in a trained model, the accuracy of \emph{Running} is very low. Most \emph{Running} videos are classified as \emph{Walking}. Then, we analyzed the difference between these two actions and found that the action of \emph{Running} was faster than that of \emph{Walking}. Therefore, we added ``the action is fast'' as a feature to \emph{Running}. During the testing phase, if the video has the feature of ``the action is fast,'' then the probability of \emph{Running} increases. 

One of the advantages of the proposed model is that it can be continuously optimized by modifying the text features of actions without retraining the model.
For example, during the testing phase, the videos of \emph{StillRings} were incorrectly classified as \emph{HandstandWalking}. We found that \emph{HandstandWalking} had only one feature, which is ``A person stands upside down,'' whereas people in \emph{StillRings} also have this feature.
We then added a new feature to \emph{HandstandWalking}, which is ``A person walks with his hands on the ground.''
Thereafter, most \emph{StillRings} videos were correctly classified.
In this way, the accuracy of the model can be continuously improved.

The major contributions of this study are as follows:

\begin{enumerate}
	\item We propose VTMM, which calculates the matching degree between the video and the text feature.

	\item The model can be used independently to predict the class of a video. This requires manual annotation of text features for each action class.

	\item The model can be appended to any other model to improve the accuracy of the model; features are added only to the classes that needs to be optimized without modifying other classes. This method requires very little labor.

	\item We continuously optimized the proposed model by modifying the text features of each action, which does not need to retrain the model. The optimization effect can be observed immediately after the modification.
\end{enumerate}

The remainder of this paper is organized as follows. 
Section 2 presents related works. 
Section 3 introduces the architecture of the proposed model and its optimization methods.
Section 4 presents the experimental results and discussions. 
Section 5 presents the conclusion.

\section{Related Work}\label{sec2}
This section review related works on action recognition, zero-shot learning, zero-shot action recognition, and visual question answering.

\subsection{Action Recognition}

Previously, action recognition artificially crafted features to obtain the movement trajectory of people \cite{iDT,wang2013dense}.
In recent years, owing to the development of deep learning, action recognition has significantly developed. Recognition accuracy is improving and the models are becoming increasingly complex. More people are using deep learning algorithms for action recognition. There are two main types of action recognition methods.
The first method uses the image data of videos to recognize actions \cite{two-stream,yang2020temporal,li2020tea,majd2019motion,elharrouss2021combined}. The advantage of this method is that the amount of data is large and contains considerable information. ImgAud2Vid \cite{gao2020listen} also added audio data to improve recognition accuracy. Nevertheless, processing the data requires huge computer memory and computing speed. 
The second method uses human skeleton data from videos to recognize actions \cite{shi2019skeleton,cheng2020skeleton,li2019actional,liu2020disentangling}. The advantage of this method is that the amount of data is small, there is little redundant data, and it is easy to process. However, at times there is so little data that it is difficult to recognize the action. 
Nonetheless, both methods have advantages and disadvantages, and choosing a method depends on the situation.
\cite{franco2020multimodal,my_paper2} used RGB image data and skeleton data. Because these models can process more data, their accuracies are higher.

\begin{figure*}[tb]
	\centering
	  \includegraphics[width=0.8\linewidth]{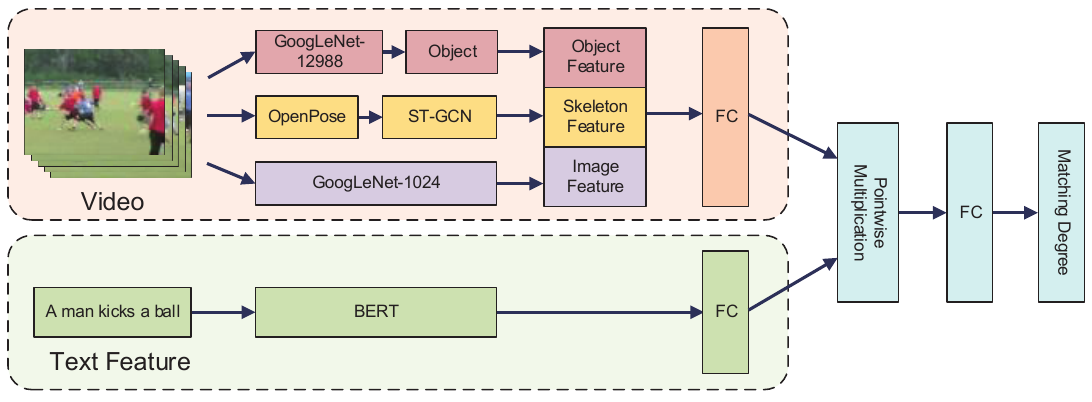}
	 \caption{ Zero-shot action recognition. }
	  \label{zsar_fig}
 \end{figure*}

\subsection{Zero-Shot Learning}
The accuracy of deep learning has significantly improved owing to algorithm improvements and the increase in the size of datasets; the larger the dataset, the more information it contains and the more the model can learn. Although a larger dataset can improve the recognition accuracy of the model, the attendant problems persist. The main problem is that annotating large datasets incurs huge labor costs, making it difficult for an ordinary person to annotate enough data, and the only choice is to give up. 

To solve this problem, zero-shot learning is the most preferred solution. With zero-shot learning, a model can recognize actions, even those that the model has never encountered before. However, the model should know the features of the actions. With the traditional classification model, data and class are input into the model, and the model learns the relationship between them. In contrast, zero-shot learning outputs the feature of the class instead of the class. In early zero-shot learning \cite{awa,lampert2013attribute}, the input and output of the model was the image data and the object features that are manually annotated, respectively, such as color and shape. Using this data, we can train a model to learn mapping between the image data and the features. When unseen images are input, the model infers the features of the images. Then, the output features can be compared with the features of all classes to identify the most similar class. Finally, the most similar class is considered as the classification result.

However, this method requires manual annotation of each class, which costs labor, and it is not easy for humans to identify and describe all the features. Therefore, in recent years, the word embedding of the class name as the feature has been used \cite{devise,kodirov2017semantic,kong2022learning}. In this method, the labels are first transformed into the semantic space. Some commonly used transform models, which transform words into vectors, are word2vec \cite{mikolov2013efficient} and GloVe \cite{glove}. 
If a label contains multiple words, the average word embedding of these words represents the word embedding of this label. Then, the vector is used as the feature of the class. The data is considered as the input and the transformed word embedding is used as the output for model training. If unseen data is input, the model transforms the data into the predicted word embedding. Finally, the cosine similarity between the predicted word embedding and word embeddings of all classes is calculated. The most similar class is determined as the result of classification. The advantage of this method is that it saves feature annotation labor costs.

For zero-shot learning, the images need to be transformed into the semantic space; therefore, the accuracy of the word embedding affects the performance of the model. If the word embeddings have errors, the results will also have errors. To solve this problem, the ensemble network \cite{ye2019progressive} used multiple word embeddings and calculated the average value as the result, which improved the robustness of the model.
Xu et al. \cite{xu2022vgse} used compact watershed segmentation to divide the image into multiple small images, which was put into the CNN to cluster the output into multiple classes. They treated each class as a feature. In the testing phase, an unseen image was input, and the model recognized the image features. These features were used to infer the class of the image.
Huang et al. \cite{huang2022robust} proposed a zero-shot object detection framework that contains two parts: (i) Intra-class semantic diverging component. It can prevent miss-classifying the real unseen objects as image backgrounds. 
(ii) Inter-class structure preserving component. It is used to avoid the synthesized features too scattered to mix up the inter-class and foreground-background relationship. Moreover, it is the first study to carry out zero-shot object detection in remote sensing imagery.

\subsection{Zero-Shot Action Recognition}
In recent years, the emerging problem in action recognition is that datasets are becoming increasingly larger. Researchers are applying zero-shot learning to action recognition to solve this problem, which is called zero-shot action recognition. 
The process of recognition is shown in Figure \ref{zsar_fig}. With this method, we can recognize an action without seeing it; however, the features of the action are needed. 
For example, we saw the video of \emph{Walking}, but not the video of \emph{Running}. However, we know that the motion trajectory of \emph{Running} is similar to that of \emph{Walking}, and that the action speed of \emph{Running} is more than that of \emph{Walking}. Based on this information, we can recognize the action of \emph{Running} without having seen the \emph{Running} video before.

Action2vec \cite{hahn2019action2vec} optimized a video processing method by dividing a video into multiple small segments and using C3D \cite{c3d} to extract features of each segment. Then, they used long short-term memory (LSTM) \cite{LSTM} to process these segment features, which makes the model better understand the relationship between the front and back content of the video.
Liu et al. \cite{liu2019generalized} used the embedding of the action name to calculate the similarity between the seen and unseen classes. They put the video into the model and calculated the probability that the video belonged to each seen class. Finally, the model was able to calculate the probability that the video belonged to each unseen class.
Instead of transforming the video into a semantic space, Kerrigan et al. \cite{kerrigan2021reformulating} transformed the video and labels to vectors and multiplied the vectors of the video and label. Then, they put the product into a fully connected network, and the final output represented the matching degrees between the video and labels.
Similarly, VD-ZSAR \cite{xing2021ventral} transformed the video and labels into vectors, which in turn were transformed into a new visual-semantic joint embedding space to identify the most appropriate label for the video in the new space.

\subsection{Visual Question Answering}
Previously, question answering (Q\&A) systems were mainly text-based Q\&A  \cite{hirschman2001natural,choi2018quac}. The user asked a question, put the question into the model, and the model answered the question. Then, researchers began to study the image-based question answering system, which is called visual question answering (VQA). An image and a question about the image are input into the model, and the model outputs the answer. 
Antol et al. \cite{antol2015vqa} first proposed the VQA. The authors only used a simple network structure but achieved good results. This model has two inputs, one is the image and the other is the question. The authors used a trained VGGNet \cite{simonyan2014very} to extract image features and LSTM to process the question. Furthermore, they used pointwise multiplication to fuse the two features and put them into a 2-layer fully connected network. Finally, the model output the answer.

Dancette et al. \cite{dancette2021beyond} believe that the judgment of the current VQA models is mainly based on the question, and not the contents of the picture. For example, when the question is ``What color is the sky,’’ the model is likely to answer the question based on previous experiences, rather than looking at pictures and analyzing the answers. The authors built a dataset of input-output variables and applied a rule-mining algorithm, which allowed the model to better understand the relationship between the question and the image.
Anderson et al. \cite{anderson2018bottom} divided the image into many small images and put the small images into the model to calculate the results. They calculated the average of all the results as the final result.

\section{Method}\label{sec3}

This section introduces our video-text matching model (VTMM) and its optimization method.

\subsection{Video-Text Matching Model}
\begin{figure*}[tb]
	\centering
	\includegraphics[width=0.8\linewidth]{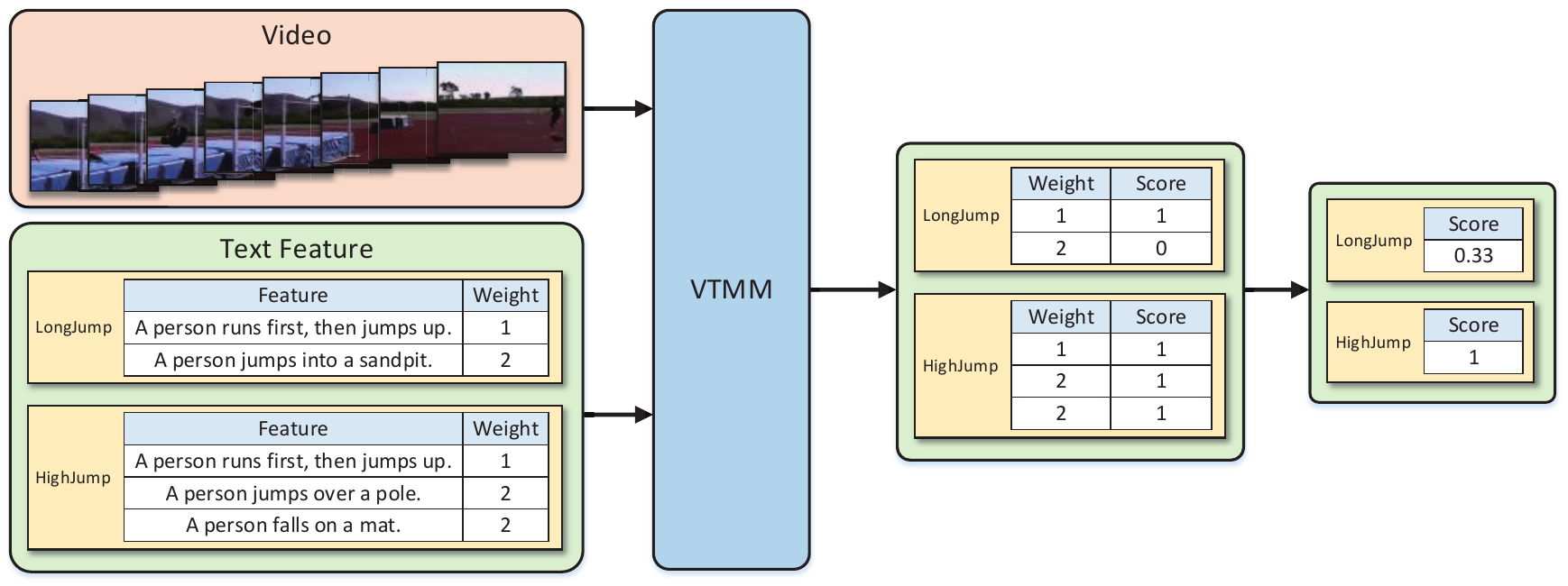}
	\caption{Overview of the proposed video-text matching model (VTMM). Our model has two inputs: a video and a text feature. The video feature has three parts: (i) objects appearing in the video, (ii) human skeleton motion data, and (iii) visual features of the video. The model calculates the matching degree between a video and a text feature}
	\label{our_model_fig}
\end{figure*}

The proposed model, as shown in Figure \ref{our_model_fig}, has two inputs: a video and a sentence, and determines whether a video and a sentence match. If the video matches the sentence, the model will output a high value; else, the model will output a low value. Although the proposed model is similar to a VQA model, the output of VQA can be a number, true, false, or other content, whereas the proposed model only output the matching degree or value.
 
For prediction, describing features with text is required; the matched features are used to predict the class of the actions.
For example, we annotate several features using text for the action \emph{Billiards}, such as ``A person is standing at the table with a stick in his hand’’ and ``There are many balls on the table.’’
Additionally, we also annotate some common features, such as \emph{indoor}, \emph{action is slow}, and \emph{standing}.
We can label multiple actions with text features and predict actions.
Based on the annotated text features and test video, our model can determine which features the video matches. 
Based on the features, which class the video belongs to can be inferred.
When a video matches a feature, it does not indicate that the video definitely belongs to the class associated with the feature, but increases the probability that the video belongs to that class.
For example, if the model detects that a person is in the kitchen, he is probably cooking.
Although he can dance, play football, or do other things in the kitchen: these are less likely.

\subsection{Features of Video}
For the generation of video features, there are several options. Commonly used methods are \cite{c3d,two-stream,i3d}.
Because we need to manually annotate the features to make the model easier to generalize, the size of the feature must be small. Finally, three features are chosen: (i) objects appearing in the video; (ii) human skeleton motion data; and (iii) visual features of the video. We concatenate these three feature vectors together as the video feature to retain a variety of information. Furthermore, the feature size is 2224 , which is not significantly large. Our model empirically works well with this size.

\subsubsection{Object Features}
To extract object features, we choose an image classifier,  GoogLeNet-12988 \cite{imagenet_shuffle}, rather than object detections.
Although object detectors can obtain the location of the object, they have few classes (usually 80 classes). In many cases, this number is insufficient for action recognition. In contrast, image classifiers can identify objects in a large number of classes (usually 1,000 classes).
GoogLeNet-12988 \cite{imagenet_shuffle} can recognize 12,988 different objects, which meets our needs. 

Furthermore, this model has so many classes that some label words are not present in the word embedding model. To solve this problem, the ImageNet \cite{deng2009imagenet} dataset is employed. This dataset contains the parent-child relationships of various classes. If the word of a class is so rare that it is not found in the word embedding model, and hence, its parent class will replace it. For example, as \emph{openbill} is not present in the word embedding model, we cannot transform it to the semantic space. Therefore, we look for this word in the parent-child relationships, and its parent class is \emph{stork}, which is present in the word embedding model. Then \emph{openbill} is replaced by \emph{stork}. If its parent class was also not found in the word embedding model, we would continue to search until a suitable word is found.

The GoogLeNet-12988 \cite{imagenet_shuffle} model can detect many objects, whereas many objects have low probabilities. Therefore, we sort all objects by probability and select the objects with the top-4 highest probabilities. 
Each object is transformed into a 300-dimensional vector with GloVe \cite{glove}.
These four vectors is used as features for the object detection part of the video.

\subsubsection{Human Skeleton Motion Features}
The most important information in action recognition is the motion of the person. To improve the generalizability of the model and reduce the dimensions of the data, we choose the skeleton data as the human motion data. We first use OpenPose \cite{openpose} to identify the human skeleton in each frame of the video. 
Then, the skeleton data is put into the ST-GCN \cite{stgcn} model to generate the motion features of the video. ST-GCN is an action recognition model. The input and output data of this model is the human skeleton data and the class of the action, respectively. To obtain the feature data of the motion, we remove the last layer of this model. 
The output of the penultimate layer is considered as the feature of the skeleton motion data. The size of the feature is 256.

\subsubsection{Visual Features}
The visual features of the video are also important. For example, the background or the brightness of the video provides information for recognizing the action. For example, if the background of a video is the kitchen, the person in the video is most likely to be cooking. To obtain the visual features of the video, we employ the image encoder of contrastive language-image pre-training (CLIP) \cite{clip}. 
In the multimodal field, CLIP has a good performance in extracting features.
After experiments, adding the image encoder of CLIP to our model can significantly improve the accuracy.
This encoder can convert an image into a vector, which can be used as the feature of the image.
We put each frame of the video into the encoder, and each frame is transformed into a vector of size 768. To reduce the amount of data, we average all the vectors as:
\begin{equation}\label{feature_eq1}
V  = \frac{1}{n} \sum_{i=1}^{n} V_{i} ,
\end{equation}
where $V_i$ is the feature of the $i$-th frame.
Finally, we obtain one vector, which is considered as the visual feature of the video. 
Each video in the dataset is only a few seconds, which is very short. This results in little difference in all frames. Therefore, the average of the features of all frames can represent this video.

\subsection{Description of Video}
For our model to be able to determine whether the video and text features match, the corresponding text features for each video are also necessary. We choose two types of text features.

 \subsubsection{Long Text Description}
 We select the VATEX \cite{vatex} dataset comprising 25,991 videos and 259,910 English captions to train the model. The description of each class of action can be one or multiple sentences. From experience, if we can describe an action from multiple perspectives, the accuracy of the action recognition will be higher. Each sentence has a weight that represents the importance of this sentence. 
If the weight is not set, the default weight is 1. The higher the weight, the more important the sentence is, and the easier it is to use this sentence to recognize the action. For example, ``A person is standing'' is so common that it is hard to tell what the person is doing.
All we know is that he is standing, and what his upper body is doing or what he is holding is unknown. He may be playing baseball or shooting arrows.
If the weight of this sentence is too high, it may increase the probability of other classes. If the sentence is ``A man is holding a stick in his hand and hits a ball.'' From this description, we can easily predict the class of the action. While we do not know what this action is, it is possible to limit the selection to a very small number.

 \subsubsection{Short Common Features}
To provide more information of actions, short features are also required.
We choose Kinetics-400 to pretrain our model, which has 400 classes of actions. Owing to the labor costs, we cannot annotate features individually for each video. Therefore, we find the common features of each class and provide these features to all videos in the class. The common features are the features that often appear in videos. 
After analyzing the features of all the classes, 20 common features are selected. 
These features can be divided into three types.
The first is the scene, such as \emph{indoor}, \emph{on the grass}, \emph{in the stadium}.
The second is the number of people, such as \emph{only one person}, \emph{so many people}.
The third is the speed of the action, such as \emph{action is fast}, \emph{action is slow}.
We use common features as a supplement to long features, which is unnecessary. 
When describing an action, we just say what the person is doing. However, the background and speed of the action are also important information, which are easily overlooked.
Therefore, we specifically list these features.
If we do not use the common feature and input the content as a sentence, the same effect will be achieved.
These features are not constant; if there are other suitable features, we can add them if they can be described in words.

 \subsubsection{Transform Texts to Vectors}
We have several options for transforming text features into vectors. A common method is to transform all words into word embeddings, and take the average of all embeddings as the embedding of the sentence. Another method is to use a trained LSTM, recurrent neural network (RNN), or bidirectional encoder representations from transformers (BERT) \cite{bert} to transform sentences directly into vectors as sentence embeddings. 
Additionally, the text encoder of CLIP \cite{clip} can convert text into vectors. Of these methods, CLIP has the highest accuracy and best effect on cross-modal. Therefore, we choose CLIP to transform texts into vectors. If the text content is put into the model, it will output a vector of length 768, which represents the feature of the text content.

\subsection{Pre-training of the Model}
Our model has two inputs; one is the feature of video, which is a vector of length 2224., and the other is the feature of the text description, which is a vector of length 768. 
Both vectors are inputted into two fully connected networks individually, and the size of both output vectors becomes 1500. We multiply the two vectors pointwise and put the result into a 3-layer fully connected network. Subsequently, the result is obtained. The result represents the matching degree of the video and text description, whose value is between zero and one. 
The threshold is set to 0.5, and if the matching degree is greater than this value, we consider the video and text description to be matched. 
In contrast, if the matching degree is less than the threshold, the video and text description are considered not to match.
For pre-training, a training set is needed. 
In the VATEX dataset, each video has 10 sentences describing the video. 
The first step is to generate positive samples.
We combine a video and its 10 description sentences separately as 10 samples. 
Each video is processed similarly and the result of each sample is set to one. 
These data is considered as positive samples. 
The second step is to generate negative samples.
We randomly pick a video and a description text in all but the class of this video. 
This description text and video is considered as a sample and set the result to zero. 
The generated negative samples has the same number of positive samples.

\begin{figure*}[tb]
	\center
	  \includegraphics[width=0.8\linewidth]{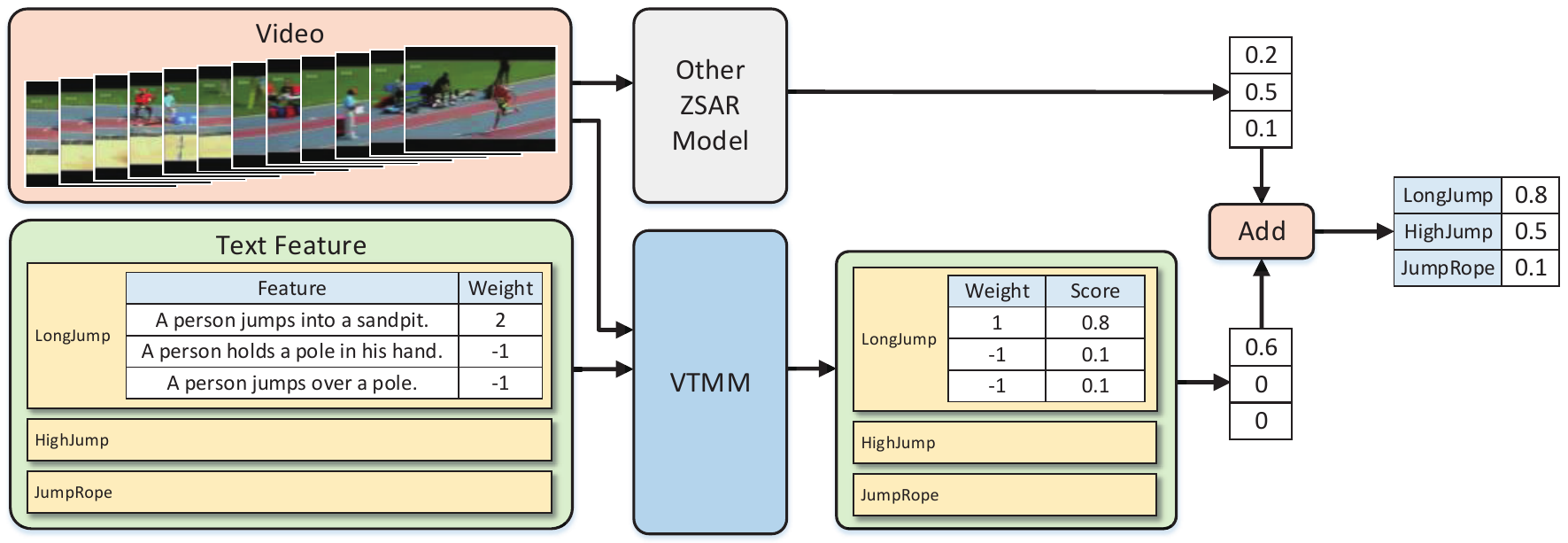}
	 \caption{Use VTMM alone for classification. We annotate text features for each class}
	\label{alone_fig}
\end{figure*}

\subsection{Use the Model Alone}
As shown in Figure \ref{alone_fig}, the model can be used alone for action recognition. All classes need to be annotated with text features, and each class has at least one feature. We can use common phrases or long sentences as features. To improve the accuracy of classification, more features are needed. We add a weight to each feature based on the importance of the feature. The default value for this weight is 1. For a video that needs to be classified, this video and all the annotated text features are input into the model. The model will output the matching degrees between the video and text features.
Each class of actions has several features. We classify all features using the class of the action they belong to. All the features of an action are put together. Then, we can get the matching degree and weight of each feature in this action. We use the weighted average method to calculate the matching degree between the video and action class:
\begin{equation}\label{alone_eq1}
s_{p} = \frac{\sum_{i=1}^{k}w_{pi} s_{pi} }{\sum_{i=1}^{k}w_{pi} } , 
\end{equation}
\begin{equation}\label{alone_eq2}
s_{n} = \frac{\sum_{j=1}^{m}w_{nj} s_{nj} }{\sum_{j=1}^{m}w_{nj} } ,
\end{equation}
\begin{equation}\label{alone_eq3}
s = s_{p} + s_{n},
\end{equation}
where $w_{pi}$ and $s_{pi}$ are the positive weights and corresponding scores, respectively. $w_{nj}$ and $s_{nj}$ are the negative weights and corresponding scores, respectively. $s_{p}$ and $s_{n}$ are the weighted average of all scores, which have positive and negative weights, respectively. $s$ is the matching degree between the video and action class.
In this way, we can obtain the matching degrees of this video and all actions. 
Finally, the class with the highest matching degree is chosen as the result for this video.

 \begin{figure*}[tb]
	\center
	\includegraphics[width=0.9\linewidth]{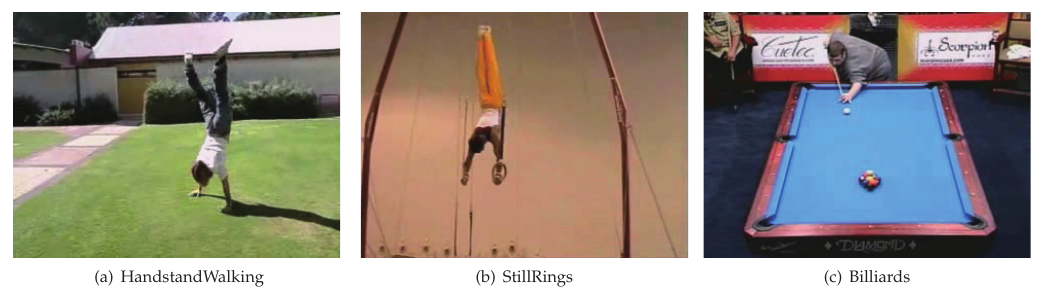}
	\caption{Append VTMM to another model. We only annotate text features for the classes that need to be optimized}
	\label{append_fig}
\end{figure*}

\subsection{Append to Other Models}
As shown in Figure \ref{append_fig}, our model can be add to any other model. The accuracy of the original model can be improved without retraining or modifying the original model. This method can annotate features for only one action, instead of annotating features for all actions. With this method, certain actions can be optimized in a targeted manner. For example, we found that the action accuracy of a certain class is low, and the videos of this class are wrongly classified into various classes. If the accuracy of one class is low, there are two methods to optimize: (i) If the videos of this class are wrongly classified into various classes, we analyze the features of this action and describe these features in texts. (ii) If the videos of this class are wrongly classified into the same class, we analyze the difference between the two actions and describe them in texts. For example, if most of the \emph{Running} videos are wrongly classified as \emph{Walking}, and the difference between them is that the \emph{Running} action is faster than \emph{Walking}, we describe the \emph{Running} feature as ``The action is fast.'' Subsequently, the description texts are used as the features of \emph{Running}. In the testing phase, we put the video and features of all actions into the model to obtain the matching degrees between the video and features. We classify all features by the actions they belong to and then use the weighted average method to calculate the matching degrees between the video and annotated actions. 
We transform the matching degree into a correction value to modify the results of the original model:
\begin{equation}\label{append_eq1}
S_{\mathrm{final}} = S_{\mathrm{origin}} + \lambda \cdot S_{\mathrm{VTMM}},
\end{equation}
where $S_{origin}$ and $S_{VTMM}$ are the results of the original model and the proposed model, respectively, and $\lambda$ is the correction factor.
The correction factor is set empirically. If this correction factor is too high, it is easy to classify videos of other classes into this class. In contrast, if the correction factor is too low, the effect of the correction will be poor. In general, $\lambda$ is set to 1. Finally, we obtain a new result. The class with the highest score is chosen as the action class of the video.

\subsection{Instruction and Optimization}
One of the advantages of the proposed model is that we can optimize it repeatedly without retraining it. For other models, if the accuracy is low, the dataset or parameters have to be adjusted. Then, the model is retrained, which consumes labor and time. Furthermore, the retrained model is not necessarily better. Moreover, a user who does not know how to train the model using the software cannot optimize the model. The errors of the model will persist until a professional modifies the model. Conversely, by using the proposed, users can analyze the results and continuously optimize the model by adding a few text features to improve its accuracy, without retraining it. After modifying the features, the model accuracy improves immediately.

For example, during the testing phase, the videos of \emph{StillRings} were incorrectly classified as \emph{HandstandWalking}. Then, after analyzing the cause of the error, we found that among the features, \emph{HandstandWalking} has only one feature, which is ``A person stands upside down.'' Whereas people in \emph{StillRings} also have this feature, which was particularly noticeable. Conversely, \emph{StillRings}' own feature was ``A person holds rings.'' In the video, this feature is less noticeable considering the ring is too small to be noticed, so the matching degree of this feature was low. We input the videos of \emph{StillRings} to the model, and the model only detected the feature of \emph{HandstandWalking} instead of its own features. Therefore, the videos of \emph{StillRings} were wrongly classified as \emph{HandstandWalking}. By adding the differences between the two actions to the model, this error can be corrected. After analysis, we found that although the people in these two actions are upside down, the person in \emph{HandstandWalking} is walking, and the person in \emph{StillRings} is always in one place. Therefore, we added a new feature to \emph{HandstandWalking}, which is ``A person walks with his hands on the ground.'' After adding this feature, because \emph{StillRings} did not match this feature, most of \emph{StillRings} videos were correctly classified, which can help continuously improve the accuracy of the model. The maximum accuracy of zero shot action recognition should reach the accuracy of VTMM.

\section{Experiments}\label{sec4}

This section describes the experiments conducted on UCF101 \cite{ucf101} and HMDB51 \cite{hmdb} to evaluate the performance of our model.

\subsection{Dataset}
The VATEX \cite{vatex} dataset is chosen to pre-train our model. This dataset had 25,991 videos and 259,910 English captions. Each video was described by 10 captions. Each caption described all or part of the video. VATEX videos are part of Kinetics-600 \cite{kinetics600}. We only had the Kinetics-400 \cite{kinetics} dataset, which is part of the Kinetics-600. Therefore, we analyzed the same parts of the Kinetics-400 and VATEX and used 13,568 VATEX videos to train our model. We analyzed the videos and manually listed 20 common features and annotated these features for each action. The features of one class action are shared with all the videos of this class. In this way, all videos will be annotated with features.
UCF101 and HMDB51 are chosen to evaluate our model. UCF101 had 13,320 videos and 101 actions. The 101 classes are divided into 51 seen and 50 unseen classes. HMDB51 had 7,000 videos and 51 actions. The 51 classes are divided into 26 seen and 25 unseen classes. 
The division of the data set is shown in Table \ref{dataset_table}. Videos in the test set did not have text features. Therefore, we manually annotated features for each class of action. We annotated 1–5 sentences as features for each action class and selected appropriate features to annotate each action among 20 short common features.

\begin{table}[tb]
	\begin{center}
		\caption{ The information of the dataset.}
          \label{dataset_table}
		\begin{tabular}{@{}cccc@{}}
			 \toprule
			 Phase         & Dataset& Class (seen/unseen)& Videos \\
			\midrule
    		 Pre-train 	  & VATEX & 381/0             & 13,568 \\
			 Test         & UCF101& 0/50              & 6,687\\
			Test 	     & HMDB51& 0/25              & 3,488\\
			 \bottomrule
		\end{tabular}
	\end{center}
\end{table}

 \begin{table*}[tb]
\begin{center}
\caption{Accuracy of Each Model.\label{result_table}}
\begin{tabular}{cccc}
\toprule
Model & Reference & UCF101 & HMDB51 \\
\midrule
DAP 		& CVPR2009 \cite{awa} 						& 14.3 & N/A 	\\
IAP 		& CVPR2009 \cite{awa} 						& 12.8 & N/A 	\\
ESZSL 		& ICML2015 \cite{romera2015embarrassingly} 	& 14.5 	& 17.6 	\\
ZSECOC		& CVPR2017 \cite{qin2017zero}   			& 15.1	& 22.6 	\\
UR			& CVPR2018 \cite{zhu2018towards}   			& 17.5 	& 24.4 	\\
Two-Stream GCN	& AAAI2019 \cite{tsgcn} 		        & 33.4 	& 21.9 	\\
CEWGAN		 & CVPR2019 \cite{mandal2019out} 			 & 26.9 	 & 30.2 \\
BD-GAN		 & Neurocomputing2020 \cite{mishra2020zero} 	 & 18.7 	 & 23.9 \\
VDARN		 & Ad Hoc Networks2021 \cite{su2021vdarn} 	 & 26.4 	 & 21.6 	 \\
ER		   & ICCV2021 \cite{chen2021elab} 		       & 51.8 	 & 35.3 	 \\
LUPI		 & ACCV2022 \cite{gao2022learning} 			 & 52.6 	 & 38.8 	 \\
ResT\_101		 & CVPR2022 \cite{lin2022cross} 			 & 58.7 	 & 41.1 	 \\
VTMM (original)        & Ours					       & 60.5 	 & 48.7 	 \\
VTMM (optimized)	 	 & Ours	       & \textbf{ 78.1 }  &  \textbf{ 59.4 } \\
ER+VTMM (optimized)	   & Ours	       & \textbf{ 81.3 }  &  \textbf{ 60.5 } \\
 \bottomrule
\end{tabular}
\end{center}
\end{table*}

\subsection{Pre-train the Model}
 \label{section:pre_train}
 First, we constructed a pre-training dataset for our model. 
The dataset is the part of VATEX, which comprises 381 action classes, 13,568 videos, and 13,5680 captions as the pre-training dataset. 
We selected suitable features from 20 short common features and manually annotated each action. 
Then, the video were transformed into features with 3 parts. 
(i) OpenPose detected the human skeleton data in the video and the skeleton data was put into ST-GCN. 
We removed the last layer of ST-GCN and used the output of the penultimate layer as the feature data of human action. 
(ii) The GoogLeNet-12988 \cite{imagenet_shuffle} detected the objects that appeared in each frame of the video and calculated the average of the results. The objects with the top-4 highest scores were selected as the objects appearing in the video and transformed the names of the objects into vectors using GloVe \cite{glove}. These four vectors were concatenated as the object features of the video. 
(iii) We input each frame of the video into the image encoder of CLIP \cite{clip} and used the output as the visual feature of the frame. Finally, the average of the features of each frame is the visual feature of the video.
After creating the video features, the text encoder of CLIP transformed the VATEX text and our annotated common features into vectors. These vectors were the text features of videos.
The video and text features are the two input data of our model to pre-train the model. We trained it for 2000 epochs with SGD, with learning rate, dropout, and batch size of 0.5, 0.5, and 1024, respectively. After training the model, we evaluated the accuracy of the model. The results are shown in Table \ref{VTMM_acc_table}.

 \begin{table}[tb]
	\begin{center}
		\caption{Accuracy of VTMM on VATEX.}\label{VTMM_acc_table}
		\begin{tabular}{@{}cc@{}}
			\toprule
			Long Sentence Feature & Short Common Feature\\
			\midrule
			98.2 	& 99.8	\\
			\bottomrule
		\end{tabular}
	\end{center}
\end{table}

\subsection{Model Evaluation}

\subsubsection{Use the Model Alone}
 \label{section:use_alone}
 First, we annotated each action of UCF101 and HMDB51 with one text feature. Then, we classified each video based on these annotated features. The results are shown in Table \ref{result_table}. To improve the model accuracy, we modified the features, such as adding more features to the actions with low accuracies. 
The detailed process of optimization will be shown in the optimization section \ref{section:optimization}. Thereafter, we evaluated the model again. As shown in Table \ref{result_table}, the accuracy of our model was improved, which proves that our optimization is effective.

 \begin{table}[tb]
	\begin{center}
		\caption{Top-5 accuracy of appending VTMM to another model (order by improvement).}
        \label{append_result_table1}
		\begin{tabular}{@{}cccc@{}}
			\toprule
			Action& Original& After Correction& Improvement\\
			\midrule
    		Drumming 		& 0.0 	& 96.9&	96.9\\
			HeadMassage 	& 4.1 	& 95.9& 	91.8\\
			BasketballDunk 	& 8.4 	& 95.4& 	87.0\\
			Lunges		   & 2.4 & 84.3&	81.9\\
			BrushingTeeth 	& 20.6& 98.5&	77.9\\
			\bottomrule
		\end{tabular}
	\end{center}
\end{table}

\begin{table}[tb]
	\begin{center}
		\caption{Top-5 accuracy of appending VTMM to another model (order by accuracy).}
        \label{append_result_table2}
		\begin{tabular}{@{}cccc@{}}
			\toprule
			Action& Original& After Correction& Improvement\\
			\midrule
    		IceDancing 		& 100.0 	& 100.0&	0.0\\
			PlayingSitar 	& 100.0 	& 100.0& 	0.0\\
			Diving 	       & 78.0 	& 100.0& 	22.0\\
			Kayaking		& 91.5    & 99.3&	7.8\\
			Biking 	       & 86.6    & 99.3&	12.7\\
			\bottomrule
		\end{tabular}
	\end{center}
\end{table}

 \begin{figure*}[tb]
	\center
	\includegraphics[width=0.9\linewidth]{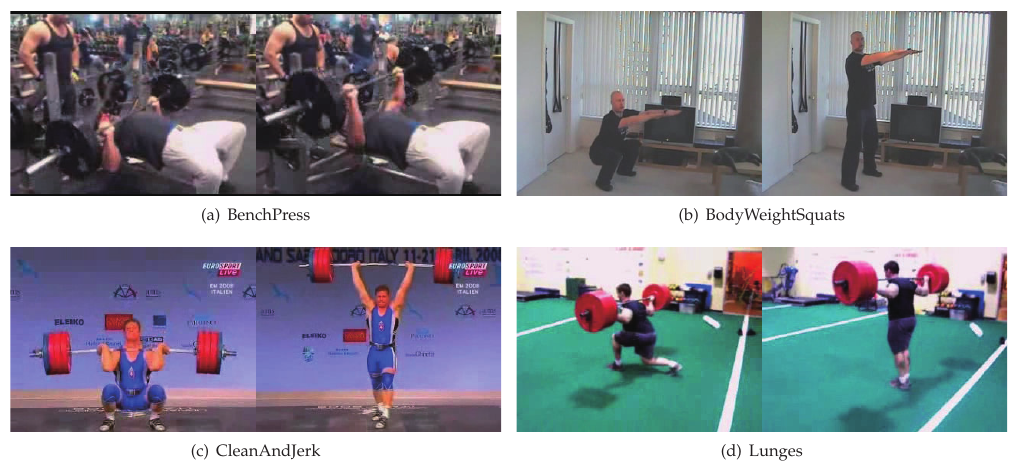}
	\caption{We input the feature---``A person stands upside down''. Three classes of actions have high scores}
	\label{false_a_fig}
\end{figure*}

 \begin{figure*}[tb]
	\center
	\includegraphics[width=0.7\linewidth]{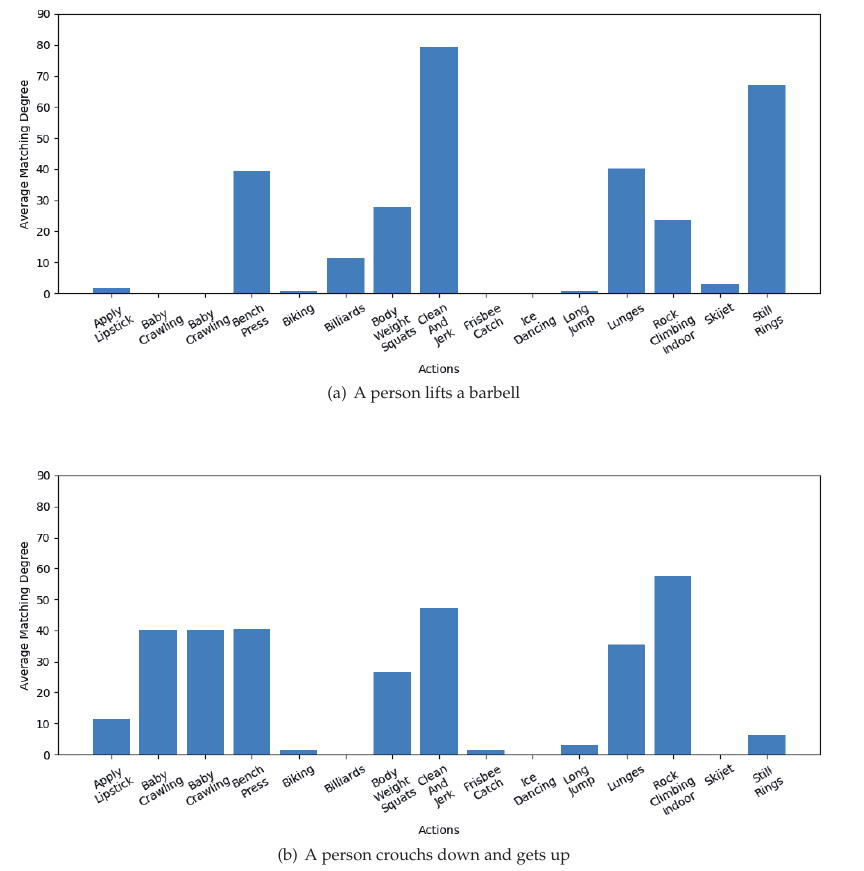}
	\caption{Some actions that are easy to be wrongly recognized}
	\label{false_c_fig}
\end{figure*}

\begin{figure*}[tb]
	\center
	\includegraphics[width=0.4\linewidth]{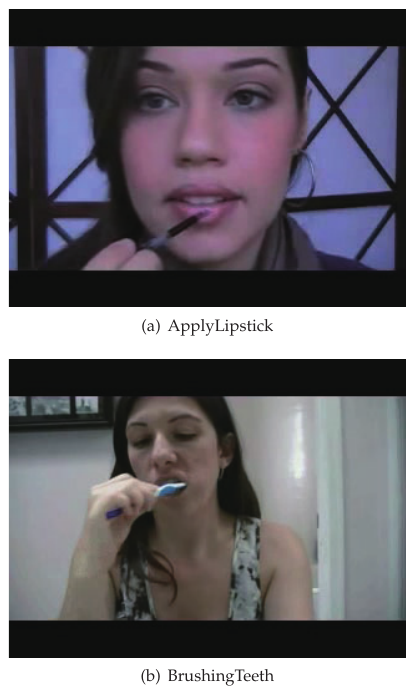}
	\caption{Matching degrees of some features}
	\label{feature_acc_fig}
\end{figure*}

 \subsubsection{Append to Other Models}
Furthermore, our model can be used to improve the accuracy of other models. We selected the elaborative rehearsal (ER) \cite{chen2021elab} as the baseline and found actions with low accuracies for optimization. 
We found videos have were misclassified and analyzed the difference between the true and wrong classes of the video. A distinction is then made into a feature, giving the action.
This process was repeated until the accuracy is difficult to improve, and the results are shown in Table \ref{result_table}.
We sorted all classes by accuracy and improvement. The top-5 classes are listed in Tables \ref{append_result_table1} and \ref{append_result_table2}.

\subsubsection{Instruction and Optimization}
  \label{section:optimization}
In this section, we present the optimization process of using the model alone. At the beginning of the experiment, we annotated each action class with one feature. We only considered the features of the actions and not the difference between the actions. After annotating the features, we evaluated our model, and the results are shown in Table \ref{result_table}. The accuracy was low at this time and we continued to add features to the classes with low accuracy to make the description more detailed. Finally, through continuous optimization, the accuracy of the model increased significantly, and the results are shown in Table \ref{result_table}. We annotated a total of 98 features, with an average of 1.96 features per action class and an average of 7.4 words per feature.

 During optimization, we found that the feature of \emph{HandstandWalking}---``A person stands upside down'' had high matching degrees with \emph{StillRings} and \emph{Billiards}. The video is shown in Figure \ref{false_a_fig}. We analyzed the difference between the three actions and annotated the new features to the actions. After optimization, the errors of the actions were reduced and the accuracy was improved.

\begin{table*}[tb]
	\begin{center}
		\caption{Features of \emph{BenchPress}, \emph{BodyWeightSquats}, \emph{CleanAndJerk} and \emph{Lunges}.}
		\label{new_feature_tab}
		\begin{tabular}{ll}
			\toprule
			Action		& Features\\
			\midrule
			\multirow{2}{*}{BenchPress} 	
            & A person lifts a barbell.\\
			& A person is lying down. \\
			\hline
			\multirow{2}{*}{BodyWeightSquats} 
            & A person crouches down and gets up.\\ 
            & A person stretches his arms forward.\\
			\hline
            \multirow{2}{*}{CleanAndJerk} 
            & A person lifts a barbell over his head.\\ 
			& A barbell is on the ground.	\\
			\hline
			\multirow{4}{*}{Lunges} 	
            & A person crouches down and gets up.\\
			& A person holds a barbell in the hand and crouches down.\\
			& A person has a barbell on his shoulders.\\		
            & A person walks with a barbell in the hands.\\	
			\bottomrule
		\end{tabular}
	\end{center}
\end{table*}

 The actions of \emph{BenchPress}, \emph{BodyWeightSquats}, \emph{CleanAndJerk}, and \emph{Lunges} were very similar, as shown in Figure \ref{false_c_fig}. Therefore, the videos of these actions were often misclassified at the beginning.
\emph{BenchPress}, \emph{CleanAndJerk}, and \emph{Lunges} have the same feature---``A person lifts a barbell.''
\emph{BodyWeightSquats}, \emph{CleanAndJerk}, and \emph{Lunges} have the same feature---``A person crouches down and gets up.''
We put these two features into the model, and the results are shown in Figure \ref{feature_acc_fig}.
There are many actions that have high matching degrees with the feature ``A person crouches down and gets up.'' In particular, \emph{BabyCrawling} and \emph{RockClimbingIndoor} are treated as squatting and getting up, when in reality, they are crawling and climbing. This can be attributed to the fact that they have similar leg motions. However, as the object features and other motion features of the two actions are easy to distinguish, they are easy to correct.
For \emph{BenchPress}, \emph{BodyWeightSquats}, \emph{CleanAndJerk}, and \emph{Lunges}, these actions are not only similar in the human motion but also in the objects appearing in the video. Therefore, we appended some new features to the actions. The features are shown in Table \ref{new_feature_tab}. By adding new features, the accuracies of these actions were improved.

 \begin{figure}[tb]
	\center
	\includegraphics[width=0.4\linewidth]{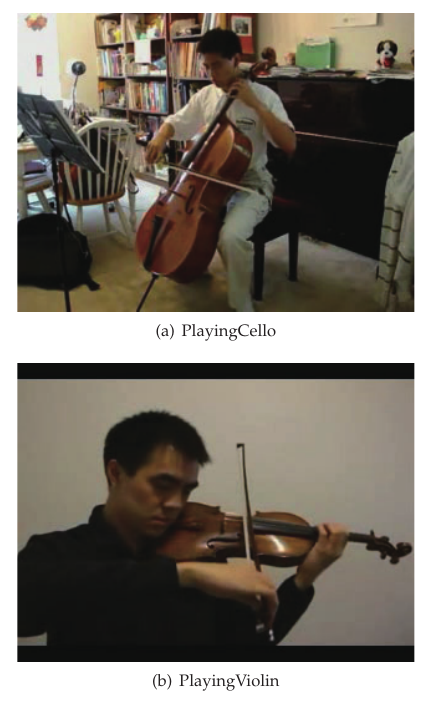}
	\caption{For similar objects and actions, it is difficult for the model to distinguish them.}
	\label{false_b_fig}
\end{figure}

 However, the errors of \emph{PlayingCello} and \emph{PlayingViolin} are difficult to correct. The video is shown in Figure \ref{false_b_fig}. 
In the field of image recognition, if the size and orientation of the object change, many models can still recognize it correctly, which is an advantage.
However, there is also a problem that is easy to appear. For example, the violin and cello look alike, but the size and orientation is different.
This can be confusing for models.
Additionally, human actions are also similar.
It is difficult for people to describe the difference of these two actions in simple sentences.
To solve this problem, we can use complex sentences to describe the subtle differences between the two actions in detail.
Moreover, it is necessary to add the data of subtle movements to the training set of the VTMM to recognize it.

\subsection{Hyperparameter Evaluation}
\subsubsection{Cross Validation}
\label{section:cross_val}
In this section, we perform cross-validation on our model.
Our VATEX dataset has 381 classes. Each time we select 50 different classes as the test set, the remaining 331 classes act as the training set.
This selection was repeated 7 times.
During the training phase, the training method is same as pre-training the model \ref{section:pre_train}.
The training set has 331 classes, which are then transformed into positive and negative samples.
In the testing phase, we evaluate the model with 3 methods:
(i) We use the remaining 50 classes of datasets to generate positive and negative samples to test the ability of the model to distinguish video and feature matching.
If one feature belongs to a video and the model output is greater than 0.5, it is considered correct.
Conversely, if one feature does not belong to a video and the model output is less than 0.5, it is considered correct.
Finally, the total accuracy of the model can be calculated.
(ii) We performed a linear evaluation on the model.
The original test set was split into training and test sets in the ratio 9:1.
The last fully connected layer of the model is fine-tuned with the new training set. 
Then, we use the new test set to test the model. 
The calculation method of accuracy is the same as (i).
(iii) Each video in the VATEX dataset has 10 sentences describing actions.

We collect all the videos of an action and put together the descriptions of these videos. 
After statistics, each action has 346.2 sentences as features on average, and each feature has 14.2 words.
We randomly pick 10 sentences from the descriptions of an action as the features.
Because each description is specific to a particular video and not an action, the descriptions lack generality.
This factor may reduce the accuracy of recognition.
The method of calculating the accuracy is the same as using the model alone \ref{section:use_alone}.
According to these methods, we experimented 7 times, and the results are shown in Table \ref{cross_val_table}.
The results show that there is little difference in the results of 7 experiments, and the model has good generalization ability.
Fine-tuning the last layer can improve the performance of the model, but the impact is small.

\begin{table}[tb]
	\begin{center}
		\caption{The accuracy of cross validation. We evaluate the model with 3 methods: (i) Evaluate whether the model can correctly calculate the matching degree of videos and features, (ii) after the fine-tuning of the last layer of model, and evaluating the model, and (iii) action recognition with features.}
        \label{cross_val_table}
		\begin{tabular}{@{}cccc@{}}
			\toprule
			Experiment ID& Original& Linear Evaluation& Action Recognition\\
			\midrule
    		1 	&  97.3  &   97.4  &  77.3 \\
			2 	&  96.4  &   97.1  &  78.4 \\
			3 	&  97.3  &   97.8  &  79.5 \\
			4	&  96.8  &   96.8  &  78.5 \\
   			5	&  97.4  &   97.7  &  78.8 \\
      		6	&  97.3  &   97.5  &  78.5 \\
         	7	&  97.5  &   98.2  &  78.9 \\
			\bottomrule
		\end{tabular}
	\end{center}
\end{table}

\subsubsection{Impact of the Feature Number}

\begin{figure*}[tb]
	\center
	\includegraphics[width=0.5\linewidth]{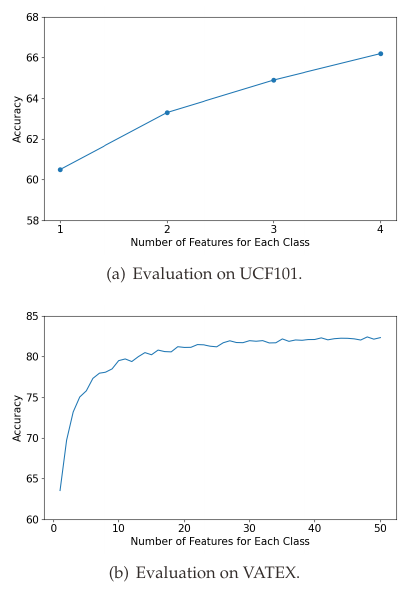}
	\caption{Accuracy of different number of features.}
	\label{feature_num_fig}
\end{figure*}

In this section, we evaluate the impact of the number of features on the recognition accuracy.
First, we chose the UCF101 dataset, used one feature to describe an action and calculated the accuracy.
Then, we describe the action in another way and use this sentence as a new feature.
Furthermore, more features are added, up to four features per action.
We did not optimize the features but simply increased the number of features.
The results are shown in Figure \ref{feature_num_fig}.
The results show that increasing the number of features can improve the accuracy of recognition, regardless of whether the new features are beneficial.

To evaluate the impact of a larger number of features, we chose the VATEX dataset.
The training method is the same as the cross-validation section \ref{section:cross_val}. We randomly selected 331 classes to train the model and used 50 classes to verify the accuracy of action recognition.
We collect all the videos of an action and put together the descriptions of all the videos. 
We randomly selected several sentences from these descriptions as features for an action.
These features are then used to classify the video and calculate the accuracy.
To reduce the error, experiments with the same number of features were carried out 20 times, and the average value was taken as the result.
Finally, we evaluate the impact of changing the number of features from 1 to 50 on the accuracy, and the results are shown in Figure \ref{feature_num_fig}.
The results show that that as the number of features increases, the accuracy of action recognition increases.

\section{Conclusions}\label{sec5}
In zero-shot action recognition, all information about the actions comes only from the word embedding of the labels, which contains very little information. Therefore, the accuracies are still low, which limits its practical application.
This study used text features instead of word embedding to solve the above-mentioned problems. We proposed VTMM, which calculates the matching degree of video and text feature. Furthermore, we created text features of test videos to provides action information. We used our model alone to predict the UCF101 and HMDB51 videos and evaluated the model. Then, we appended our model to another model and evaluated the new model. Additionally, we continuously improved the accuracy of the model by modifying the text features of actions.
The results with UCF101 and HMDB51 showed that our model achieved the best accuracy and improved the accuracies of other models.
Although manually creating features incurs some labor costs, in many situations, the costs are worth it. 
In practice, some features may not be easy to describe. For example, for two different styles of dance, although people can feel that the two types of dance are different, they may be unable to describe the difference in words. For such situations, we need to add other features in the future to better recognize the actions.
In future research, we can add a module to automatically generate text descriptions.
In this case, no labor cost will be required.


\bibliographystyle{unsrt}  
\bibliography{bibliography}

\end{document}